\documentclass[letterpaper, 10 pt, conference]{ieeeconf}  

\IEEEoverridecommandlockouts                              

\usepackage[utf8]{inputenc}
\usepackage{amsmath}
\usepackage{amsfonts}
\usepackage{amssymb}
\usepackage{graphicx}
\usepackage{epstopdf}
\usepackage{color}
\usepackage{epstopdf}
\usepackage{units}
\usepackage[caption=false]{subfig}
\usepackage[titlenumbered,ruled,linesnumbered]{algorithm2e}
\usepackage[hidelinks]{hyperref}
\usepackage[capitalise]{cleveref}

\crefname{equation}{}{} 
\crefname{section}{Sec.}{Sec.}

\newcommand{\T}{\mathrm{T}}                     

\newcommand{\mb}[1]{\mathbf{#1}}        
\newcommand{\ts}[1]{_{#1}}           

\newcommand{\ti}{n}

\graphicspath{{img/}}

\title{\LARGE \bf
Safe Controller Optimization for Quadrotors with Gaussian Processes
}

\author{Felix Berkenkamp, Angela P. Schoellig, and Andreas Krause
\thanks{Felix Berkenkamp and Andreas Krause are with the Learning \& Adaptive Systems Group (LAS), Department of Computer Science, ETH Zurich, Switzerland.
Email: \mbox{\{befelix, krausea\}@ethz.ch}}%
\thanks{Angela P. Schoellig is with the University of Toronto Institute for Aerospace Studies (UTIAS), Canada. Email: \mbox{schoellig@utias.utoronto.ca}}%
\thanks{This research was supported in part by SNSF grant {200020\_159557}, NSERC grant {RGPIN-2014-04634}, and the Connaught New Researcher Award.}%
}%

\usepackage{fancyhdr}
\newcommand{\mytitle}{\textbf{Appeared in}
\textit{Proc. of the IEEE International Conference on Robotics and Automation, 2016, pp. 493 -- 496, doi: \href{http://ieeexplore.ieee.org/document/7487170/}{10.1109/ICRA.2016.7487170} }.\\[0.7em]
\copyright 2016 IEEE. Personal use of this material is permitted. Permission from IEEE must be obtained for all other uses, in any current or future media, including reprinting/republishing this material for advertising or promotional purposes, creating new collective works, for resale or redistribution to servers or lists, or reuse of any copyrighted component of this work in other works.}
\begin{document}
\setlength{\textfloatsep}{10pt}

\maketitle
\thispagestyle{fancy}
\pagestyle{empty}

%
\begin{abstract}
One of the most fundamental problems when designing controllers for dynamic systems is the tuning of the controller parameters. Typically, a model of the system is used to obtain an initial controller, but ultimately the controller parameters must be tuned manually on the real system to achieve the best performance. To avoid this manual tuning step, methods from machine learning, such as Bayesian optimization, have been used. However, as these methods evaluate different controller parameters on the real system, safety-critical system failures may happen. In this paper, we overcome this problem by applying, for the first time, a recently developed safe optimization algorithm, \textsc{SafeOpt}, to the problem of automatic controller parameter tuning. Given an initial, low-performance controller, \textsc{SafeOpt} automatically optimizes the parameters of a control law while guaranteeing safety. It models the underlying performance measure as a Gaussian process and only explores new controller parameters whose performance lies above a safe performance threshold with high probability. Experimental results on a quadrotor vehicle indicate that the proposed method enables fast, automatic, and safe optimization of controller parameters without human intervention.
\end{abstract}


\section*{SUPPLEMENTARY MATERIAL}

A video demonstrating the proposed safe, automatic controller optimization on a quadrotor vehicle can be found at~\mbox{\url{http://tiny.cc/icra16_video}}.
A Python implementation of the algorithm is available in~\cite{BerkenkampSafeOptCode}.

\section{INTRODUCTION}

Tuning controller parameters is a challenging task, which requires significant domain knowledge and can be very time consuming.
Classical approaches to automate this process, such as the ones in~\cite{Killingsworth2006PID} and~\cite{Astrom1993Automatic}, either rely on model assumptions (e.g., linearity), which may be the very reason why the initial, model-based controller performs poorly, or require gradient approximations, which are difficult to obtain from noisy measurements.
Methods without these assumptions, such as genetic algorithms~\cite{Davidor1991Genetic}, typically require an impractical number of evaluations on the real system. Moreover, all these methods may converge to a local optimum.

\begin{figure}[t]
\vspace{1mm}
\includegraphics[scale=1]{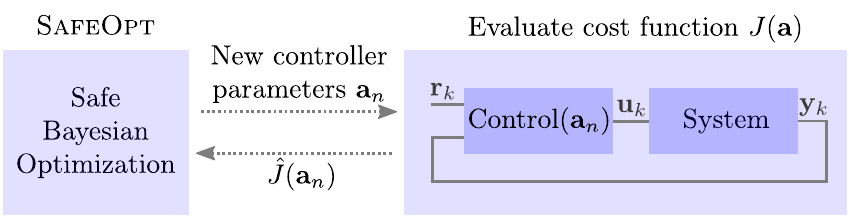}
\caption{Overview of the algorithm. The safe Bayesian optimization algorithm selects new, safe parameters at which the performance function is evaluated on the real system. Based on the noisy information gained from the experiment, the algorithm chooses a new, informative and safe evaluation point at each iteration~$\ti$. This is repeated until the optimum is found.}
\label{fig:optimization_illustration}
\vspace{-0.1cm}
\end{figure}

In this paper, we present a method to automatically tune controller parameters without requiring a model of the underlying, dynamic system or the computation of gradients.
Additionally, our approach guarantees safety during the convergence to the global optimum and only requires few experiments (see~\cref{fig:optimization_illustration}).

In general, the goal of automatic controller tuning is to find controller parameters through experiments that optimize a given performance measure. Yet the function that maps controller parameters to performance values is unknown \textit{a priori}.
Finding the global optimum of an unknown function is an impossible task. However, by making assumptions about the regularity of the unknown function, the field of Bayesian optimization has developed practical optimization algorithms (cf.~\cite{Mockus2012Bayesian}) that provably find the global optimum, while evaluating the function at only a few parameter combinations~\cite{Bull2011Convergence,Srinivas2010Gaussian}.
Another major advantage of these methods is that they explicitly model noise in the performance function evaluations.
Bayesian optimization methods often model the unknown function as a Gaussian process (GP)~\cite{Rasmussen2006Gaussian}, which can guide function evaluations to locations that are informative about the optimum of the unknown function~\cite{Mockus2012Bayesian,Jones2001Taxonomy}.

Bayesian optimization has been used in robotics to automate the process of tuning controller parameters. Examples include gait optimization of legged robots~\cite{Calandra2014Bayesian,Lizotte2007Automatic} and controller optimization for a snake-like robot~\cite{Tesch2011Using}. These papers show that Bayesian optimization reliably finds the optimal controller parameters within a few experiments.
In~\cite{Marco2016Automatic} the controller parameters of a state-feedback controller were automatically tuned using Bayesian optimization. In this work, the cost matrices in the LQR framework were used as a low-dimensional representation of parameters, which made the method applicable to higher-dimensional systems.
A comparison of different Bayesian and non-Bayesian global optimization methods can be found in~\cite{Calandra2014Bayesian}.

Despite the experimental success of Bayesian optimization methods, they have one weakness in real-world experiments. While gradient-ascent methods, such as~\cite{Killingsworth2006PID}, typically improve the controller at every iteration and thereby ensure that the resulting controllers continue to be stable, informative samples in Bayesian optimization are typically far away from the original control law to gain maximum information.
This often leads to the evaluation of unstable controllers and system failures early on in the optimization process.

In this paper, we overcome the problem of safety by using a modified version of \textsc{SafeOpt}~\cite{Sui2015Safe}, a recently developed safe Bayesian optimization algorithm that builds on the results from~\cite{Srinivas2010Gaussian} and, in addition, guarantees safety by only evaluating parameters that achieve a safe performance threshold with high probability.
The algorithm retains the desirable properties of normal Bayesian optimization, such as the ability to find the safely reachable optimum~\cite{Sui2015Safe}. The result is a safe and automatic controller tuning algorithm, which is illustrated in~\cref{fig:optimization_illustration}.

Other approaches that guarantee system safety in the presence of unmodeled dynamics make assumptions about and explicitly model the uncertainties in the nominal model. In this setting, the controllers are gradually improved by estimating the unmodeled dynamics from experimental data and recomputing the control law based on this estimate. Stability can be guaranteed by ensuring that either the controller is robustly stable for all possible models within the uncertainty specification~\cite{Berkenkamp2015Safe} or the system never leaves a safe subset of the state space~\cite{Aswani2013Provably,Akametalu2014Reachability,Moldovan2012Safe}.
Both methods require a system model and uncertainty specification to be known~\textit{a priori}, which must be accurate enough to guarantee stability.
In contrast, the method presented in this paper only requires the controller structure and initial, safe controller parameters to be known. We do not use a model of the system but model the performance function directly.


The problem of Bayesian optimization subject to constraints on the performance function was previously studied in~\cite{Gelbart2014Bayesian}.
However, they did not consider this constraint as safety-critical. As a result, evaluating parameters that do not satisfy the constraints was allowed, which would mean, in our case, evaluating unsafe parameters. In contrast, we explicitly avoid the evaluation of unsafe controller parameters. The problem of safe exploration, without the goal of optimizing the performance, was considered in~\cite{Schreiter2015Safe}. In this paper, we explore and also optimize the function within the safe region.

We demonstrate the safety and performance of \textsc{SafeOpt} experimentally on a quadrotor vehicle, for which we automatically learn optimal controller parameters without failures during the experiments.
Early results of the approach were presented in~\cite{Berkenkamp2015Safe-1}.

\section{PROBLEM STATEMENT}
\label{sec:problem_statement}

This section introduces the safe optimization problem considered in this paper.
We assume that we have a nonlinear, dynamic control law of the form
\begin{equation}
\mb{u}\ts{k} = \mb{g}(\mb{y}\ts{k}, \mb{r}\ts{k}, \mb{a}\ts{\ti}),
\label{math:control_law}
\end{equation}
which may include internal states (e.g., integrators). The control law is parameterized by controller parameters,~${ \mb{a}\ts{\ti} \in \mathcal{A} }$, at iteration $n$ in a domain ${\mathcal{A}}$. At time step~$k$, the controller maps the noisy measurements of a dynamic system,~$\mb{y}\ts{k}$, and a reference signal,~$\mb{r}\ts{k}$, to control actions,~$\mb{u}\ts{k}$.
The controller aims to achieve a desired objective, such as reference tracking.
This control objective is specified in terms of a performance measure,~${ J(\mb{a}) \colon \mathcal{A} \mapsto \mathbb{R} }$, which is evaluated on the real system and assigns higher values to controllers with better performance.
The performance measure can be anything in general, but typically depends on the control inputs and output errors of the closed-loop system (see~\cref{fig:optimization_illustration}). It is evaluated over a finite time horizon.
For example, in~\cite{Calandra2014Bayesian} the average walking speed of a bipedal robot over three experiments was used.

The goal is to automatically find the controller parameters,~$\mb{a}$, that maximize the performance measure,~$J(\mb{a})$, based on noisy evaluations of~$J$ for different controller parameters,~${ \hat{J}(\mb{a}\ts{\ti}) = J(\mb{a}\ts{\ti}) + \omega }$, where~${ \omega \sim \mathcal{N}(0, \sigma_\omega^2) }$ is zero-mean Gaussian noise.
We assume that the system is safety-critical; that is, the optimization algorithm must ensure safety when evaluating new controller parameters. We do not assume any knowledge about the model of the dynamic system, which means that the dependence of~$J$ on~$\mb{a}$ is unknown~\textit{a priori} and needs to be learned as part of the optimization routine.
Additionally, the optimization procedure must be sample-efficient; that is, only few evaluations of the performance function should be carried out in order to save time and avoid system wear.
To start the optimization procedure safely, we assume that an initial set of stabilizing controller parameters (with potentially poor performance) is available.

In this paper, we encode the safety criterion as a performance threshold,~$J_{\mathrm{min}}$, below which we do not want to fall; that is,~${J(a\ts{\ti}) \geq J_{\mathrm{min}} }$ must hold with high probability for all~$\mb{a}\ts{\ti}$ at which~$J$ is evaluated.
With this definition of safety, the resulting controllers are likely to be stable, since unstable systems typically have a significantly lower performance when considering a sufficiently long time horizon.


\section{METHODOLOGY}
\label{sec:methodology}

In this section, we review GPs and Bayesian Optimization and illustrate the theory behind \textsc{SafeOpt}.

\subsection{Gaussian Process (GP)}
\label{sec:gaussian_process}

The function~$J(\mb{a})$ in~\cref{sec:problem_statement} is unknown \textit{a priori}.
We use a nonparametric model to approximate the unknown function over its domain~$\mathcal{A}$.
In particular, we use a GP to approximate~${ J(\mb{a}) }$.

GPs are a popular choice  for nonparametric regression in  machine learning, where the goal is to find an approximation of a nonlinear map,~${J(\mb{a})\colon \mathcal{A} \mapsto \mathbb{R}}$, from an input vector~${\mb{a} \in \mathcal{A} }$ to the function value~$J(\mb{a})$. This is accomplished by assuming that function values $J(\mb{a})$, associated with different values of $\mb{a}$, are random variables and that any finite number of these random variables have a joint Gaussian distribution depending on the values of~$\mb{a}$~\cite{Rasmussen2006Gaussian}.

For the nonparametric regression, we define a prior mean function and a covariance function,~$k(\mb{a}_i, \mb{a}_j)$, which defines the covariance of any two function values,~$J(\mb{a}_i)$ and $J(\mb{a}_j)$, $i,j \in \mathbb{N}$.
The latter is also known as the kernel. In this work, the mean is assumed to be zero without loss of generality. The choice of kernel function is problem-dependent and encodes assumptions about smoothness and rate of change of the unknown function. A review of different kernels can be found in~\cite{Rasmussen2006Gaussian}. More information about the kernel used in this paper can be found in~\cref{sec:results}.

The GP framework can be used to predict the function value,~${J(\mb{a}^*)}$, at an arbitrary input,~${ \mb{a}^* \in \mathcal{A} }$,  based on a set of~$n$ past observations, ${\mathcal{D}\ts{\ti} = \{ \mb{a}_i, \hat{J}(\mb{a}_i) \}_{i=1}^n}$. We assume that observations are noisy measurements of the true function value, ${J(\mb{a})}$; that is, ${\hat{J}(\mb{a}) = J(\mb{a}) + \omega}$ with ${\omega \sim \mathcal{N}(0,\sigma_\omega^2)}$. Conditioned on the previous observations, the mean and variance of the prediction are given by
\begin{align}
\mu\ts{\ti}(\mb{a}^*) &= \mb{k}\ts{\ti}(\mb{a}^*)    (\mb{K}\ts{\ti} + \mb{I}_\ti \sigma_\omega^2)^{-1} \hat{\mb{J}}_\ti ,
\label{math:gp_prediction_mean} \\
\sigma^2\ts{\ti}(\mb{a}^*) &= k(\mb{a}^*,\mb{a}^*) - \mb{k}\ts{\ti}(\mb{a}^*) (\mb{K}\ts{\ti} + \mb{I}_\ti \sigma_\omega^2)^{-1} \mb{k}\ts{\ti}^\T(\mb{a}^*),
\label{math:gp_prediction_variance}
\end{align}
where~${\hat{\mb{J}}\ts{\ti} = \left[ \begin{matrix}
\hat{J}(\mb{a}_1),\dots,\hat{J}(\mb{a}_n)
\end{matrix} \right] ^\T}$ is the vector of observed, noisy function values,
the covariance matrix~${\mb{K}\ts{\ti} \in \mathbb{R}^{n \times n}}$ has entries ${[\mb{K}\ts{\ti}]_{(i,j)} = k(\mb{a}_i, \mb{a}_j)}$, ${i,j\in\{1,\dots,n\}}$, and
the vector
${\mb{k}\ts{\ti}(\mb{a}^*) =
\left[ \begin{matrix}
	k(\mb{a}^*,\mb{a}_1),\dots,k(\mb{a}^*,\mb{a}_n)
\end{matrix}  \right]}$
contains the covariances between the new input~$\mb{a}^*$ and the observed data points in~$\mathcal{D}\ts{\ti}$.
The identity matrix is denoted by~${ \mb{I}_\ti \in \mathbb{R}^{\ti \times \ti} }$.

\subsection{Bayesian Optimization}
\label{sec:bayesian_optimization}

Bayesian optimization aims to find the global maximum of an unknown function~\cite{Mockus2012Bayesian}. The assumption is that evaluating the function is expensive, while computational resources are cheap. This fits our problem in~\cref{sec:problem_statement}, where each evaluation of the performance function corresponds to an experiment on the real system, which takes time and causes wear.

In general, Bayesian optimization models the objective function as a random function and uses this model to determine informative sample locations. A popular approach is to model the underlying function as a GP, see~\cref{sec:gaussian_process}.
GP-based methods use the mean and variance predictions in~\cref{math:gp_prediction_mean,math:gp_prediction_variance} to compute the next sample location.
For example, \cite{Srinivas2010Gaussian} evaluates, at iteration~$n$, the parameters
\begin{equation}
\mb{a}\ts{\ti} = \underset{\mb{a} \in \mathcal{A}}{\mathrm{argmax}}~ \mu\ts{\ti-1}(\mb{a}) + \beta\ts{\ti} \sigma\ts{\ti-1}(\mb{a}),
\label{math:gp_ucb}
\end{equation}
where~${\beta\ts{\ti}}$ is a iteration-varying scalar that defines the confidence interval of the GP.
Intuitively,~\cref{math:gp_ucb} selects new evaluation points at locations where the upper bound of the confidence interval of the GP estimate is maximal.
Repeatedly evaluating the system at locations given by~\cref{math:gp_ucb} improves the mean estimate of the underlying function and decreases the uncertainty at candidate locations for the maximum, such that the global maximum is found after a finite number of iterations, cf.~\cite{Srinivas2010Gaussian}.

While~\cref{math:gp_ucb} is also an optimization problem, solving it does not require any evaluations on the real system but only uses the GP model. This corresponds to the assumption of cheap computational resources.

\subsection{Safe Bayesian Optimization}
\label{sec:safe_opt}

In this paper, we build upon the safe optimization algorithm \textsc{SafeOpt}~\cite{Sui2015Safe}.
\textsc{SafeOpt} is a Bayesian optimization algorithm, see~\cref{sec:bayesian_optimization}, which aims to maximize an unknown function by modeling it as a GP over a finite set of parameters~$\mathcal{A}$.
However, instead of optimizing the underlying function globally, it restricts itself to a safe set~${\mathcal{S} = \{ \mb{a} \in \mathcal{A} ~|~ J(\mb{a}) \geq J_{\mathrm{min}} \} }$, which only contains parameters that lead to a performance value above the safe threshold,~$J_{\mathrm{min}}$.
This safe set is not known initially, but is estimated after each function evaluation.
In our case, the initial, safe set,~$\mathcal{S}_0$, corresponds to the initial, safe controller parameters,~$\mb{a}_0$.

In this setting, the challenge is to find an evaluation strategy similar to~\cref{math:gp_ucb}, which at each iteration~$\ti$ not only aims to find the global maximum within the currently known safe set~$\mathcal{S}_\ti$ (exploitation), but also to increase the set~$\mathcal{S}_\ti$ of controllers that are known to be safe (exploration).
\textsc{SafeOpt} provides a solution to this problem~\cite{Sui2015Safe} by choosing for the next experiment the safe controller parameters about whose performance we are most uncertain. Parameters are chosen from two sets: the set of potential maximizers,~$\mathcal{M}_\ti$, whose values can lie above the current maximum according to the GP estimate, and the set of potential expanders,~$\mathcal{G}\ts{\ti}$, which can expand the set of safe controllers,~$\mathcal{S}_\ti$ (see~Fig.~\ref{fig:set_example}).

In~\cite{Sui2015Safe} these two sets were estimated using a Lipschitz constant. However, in practical applications this represents an additional tuning parameter. In the next section, we modify the algorithm in~\cite{Sui2015Safe} to use the GP's prediction directly in order to estimate these sets.


\section{MODIFIED \textsc{SafeOpt} ALGORITHM}
\label{sec:safe_opt_adapted}

\begin{figure*}[t]
\vspace{-0.2cm}
\centering
\subfloat[Initial, safe parameters.]{\includegraphics[scale=1]{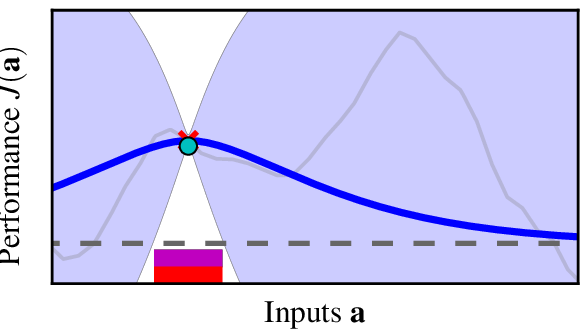} \label{fig:set_example_initial} } \hfill
\subfloat[After 5 evaluations: local maximum found.]{\includegraphics[scale=1]{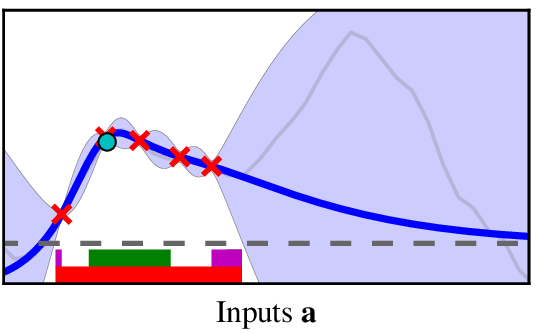} \label{fig:set_example_intermediate}} \hfill
\subfloat[After 13 evaluations: global maximum found.]{\includegraphics[scale=1]{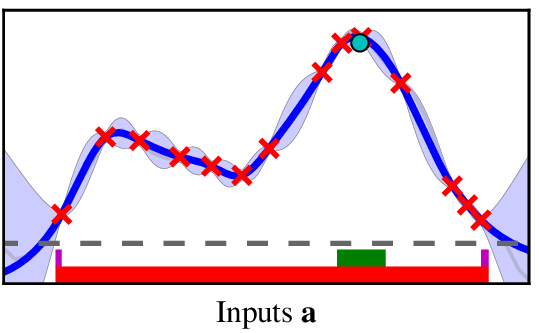} \label{fig:set_example_final}}
\caption{Optimization with modified \textsc{SafeOpt} algorithm after 1, 5 and 13 evaluations of the performance measure. Based on the mean estimate (blue) and confidence interval (light blue), the algorithm selects safe evaluation points above the safe threshold~$J_{\mathrm{min}}$ (black dashed) from the safe set~$\mathcal{S}_\ti$ (red), which are either potential maximizers~$\mathcal{M}_\ti$ (green) or expanders~$\mathcal{G}_\ti$ (magenta). It then learns about the function by drawing noisy samples from the unknown, underlying function (light gray). This way we expand the safe region (red) as much as possible, and, simultaneously, find the global optimum of the unknown function~\cref{math:optimum_estimate} (cyan ball).}
\label{fig:set_example}
\vspace{-0.5cm}
\end{figure*}

In this section, we modify \textsc{SafeOpt} from~\cite{Sui2015Safe}, to work without the specification of a Lipschitz constant. Additionally, we provide an implementation in Python~\cite{BerkenkampSafeOptCode}, which is significantly faster than a naive implementation.
In contrast to the original algorithm in~\cite{Sui2015Safe}, we estimate the sets,~$\mathcal{S}_\ti, \mathcal{G}_\ti, \textnormal{ and } \mathcal{M}_\ti$, directly from the GP. In particular, we define the upper and lower bound of the confidence interval at iteration~$\ti$ as
\begin{align}
u\ts{\ti}(\mb{a}) &= \mu\ts{\ti-1}(\mb{a}) + \beta\ts{\ti} \sigma\ts{\ti-1}(\mb{a}),
\label{math:gp_upper_bound} \\
l\ts{\ti}(\mb{a}) &= \mu\ts{\ti-1}(\mb{a}) - \beta\ts{\ti} \sigma\ts{\ti-1}(\mb{a}),
\label{math:gp_lower_bound}
\end{align}
where~${\beta_\ti \in \mathbb{R}_+}$ defines the confidence interval that we want to achieve.
For example, in our experiments we use~${ \beta_\ti = 2 }$.
Using these bounds, we define the safe set as all the parameters~$\mb{a}$ that are very likely to lead to function values above the safe performance threshold,~$J_{\mathrm{min}}$, according to the GP estimate,
\begin{equation}
\mathcal{S}\ts{\ti} = \left\{ \mb{a} \in \mathcal{A} ~|~ l\ts{\ti}(\mb{a}) \geq J_{\mathrm{min}} \right\}.
\label{math:safe_set}
\end{equation}

The set of potential maximizers contains all safe parameters that could obtain the maximum value given the high-probability bounds in~\cref{math:gp_lower_bound,math:gp_upper_bound}. It is given by the set of safe parameters for which the upper confidence interval,~$u\ts{\ti}$, is above the best, safe lower bound:
\begin{equation}
\mathcal{M}\ts{\ti} = \big\{ \mb{a} \in \mathcal{S}\ts{\ti} ~|~  u\ts{\ti}(\mb{a}) \geq \max_{\mb{a'} \in \mathcal{A}} l\ts{\ti}(\mb{a}') \big\}.
\label{math:maximizer_set}
\end{equation}

The set of potential expanders is more difficult to define without the Lipschitz constant, as it quantifies whether new parameters could be classified as safe after a new measurement.
We define an optimistic indicator function for expanders,
\begin{equation}
g\ts{\ti}(\mb{a}) = \left| \left\{ \mb{a}' \in \mathcal{A} \setminus \mathcal{S}\ts{\ti} ~|~ l_{\ti, (\mb{a}, u\ts{\ti}(\mb{a}))}(\mb{a}') \geq J_{\mathrm{min}} \right\}\right| ,
\label{math:expander_indicator}
\end{equation}
where $l_{\ti, (\mb{a}, u\ts{\ti}(\mb{a}))}$ is the lower bound of the GP,~\cref{math:gp_lower_bound}, based on past data and an artificial data point~${ (\mb{a}, u_\ti(\mb{a})) }$ with a noiseless measurement of the upper confidence bound. The function in~\cref{math:expander_indicator} counts how many previously unsafe points can be classified as safe according to~\cref{math:safe_set} assuming that we measure~$u\ts{\ti}(\mb{a})$ when evaluating~$J(\mb{a})$.
This function is positive if the new data point has a non-negligible chance to expand the safe set.
Consequently, the set of possible expanders is defined as
\begin{equation}
\mathcal{G}\ts{\ti} = \left\{ \mb{a} \in \mathcal{S}\ts{\ti} ~|~ g\ts{\ti}(\mb{a}) > 0 \right\}.
\label{math:set_expanders}
\end{equation}
For a graphical representation of the three sets, see~\cref{fig:set_example}.

As in~\cite{Sui2015Safe}, we choose new parameters at which to evaluate the performance on the real system by selecting the parameters about which we are the most uncertain from the union of the sets~${ \mathcal{G}\ts{\ti} }$ and~$\mathcal{M}\ts{\ti}$; that is, at iteration~$\ti$ we choose to evaluate the function at~$\mb{a}\ts{\ti}$,
\begin{align}
\mb{a}\ts{\ti} &= \underset{a \hspace{0.05em}\in\hspace{0.05em} \mathcal{G}\ts{\ti} \cup \mathcal{M}\ts{\ti}}{\mathrm{argmax}} ~w\ts{\ti}(\mb{a}),
\label{math:acquisition_function} \\
w\ts{\ti}(\mb{a}) &= u\ts{\ti}(\mb{a}) - l\ts{\ti}(\mb{a}).
\end{align}
This evaluation criterion has many desirable properties, including the ability to find the safely reachable optimum~\cite{Sui2015Safe}. In particular, it works well for expanding the safe set~\cite{Schreiter2015Safe}, while at the same time trading-off exploration and exploitation. For the exploration, the most uncertain parameter locations are usually on the boundary of the safe set, which results in efficient exploration.
Typical kernel functions can only classify states in the vicinity of past observations as safe, which leads to a coarse sampling of the safe parameter space.
The coarse samples already provide information about the maximizers in~$\mathcal{S}_\ti$ during exploration.
For example, all points from the set~$\mathcal{M}_5$ in~\cref{fig:set_example_intermediate} are eliminated as potential maximizers in~\cref{fig:set_example_final}, as we observe larger values during the safe exploration.
We obtain an estimate of the best currently known parameters from
\begin{equation}
\underset{\mb{a} \in \mathcal{S}\ts{\ti}}{\mathrm{argmax}} ~l\ts{\ti}(\mb{a}),
\label{math:optimum_estimate}
\end{equation}
which corresponds to the point that achieves the best lower bound on the performance.

\begin{algorithm}[t]
\SetAlgoNoLine
    \caption{Modified \textsc{SafeOpt} algorithm}
    \DontPrintSemicolon
    \label{alg:safe_opt}
    \SetKwInOut{Input}{Inputs}
    \Input{Domain $\mathcal{A}$ \newline
           Safe threshold $J_{\mathrm{min}}$ \newline
           GP prior $(k(\mb{a}_i, \mb{a}_j)$, $\sigma_\omega^2 )$ \newline
           Initial, safe controller parameters $\mb{a}_0$}
    Initialize GP with $( \mb{a}_0,~ \hat{J}(\mb{a}_0) )$ \;
    \For{$\ti=1,\dots$}{
        $\mathcal{S}\ts{\ti} \gets \left\{ \mb{a} \in \mathcal{A} ~|~ l\ts{\ti} \geq J_{\mathrm{min}} \right\}$
        \label{alg:safe_set} \;
        $\mathcal{M}\ts{\ti} \gets \left\{ \mb{a} \in \mathcal{S}\ts{\ti} ~|~  u\ts{\ti}(\mb{a}) \geq \max_{\mb{a}'} l\ts{\ti}(\mb{a}') \right\}$
        \label{alg:maximizer_set} \;
        $\mathcal{G}\ts{\ti} \gets \left\{ \mb{a} \in \mathcal{S}\ts{\ti} ~|~ g\ts{\ti}(\mb{a}) > 0 \right\}$
        \label{alg:expander_set} \;
        $\mb{a}\ts{\ti} \gets \mathrm{argmax}_{a \in \mathcal{G}\ts{\ti} \cup \mathcal{M}\ts{\ti}}~w\ts{\ti}(\mb{a})$
        \label{alg:acquisition} \;
        Obtain measurement $\hat{J}(\mb{a}\ts{\ti}) \gets J(\mb{a}\ts{\ti}) + \omega\ts{\ti}$
        \label{alg:evaluate}\;
        Update GP with $( \mb{a}\ts{\ti},~ \hat{J}(\mb{a}\ts{\ti}) )$
        \label{alg:update_gp} \;
        }
\end{algorithm}

A summary of the entire algorithm is found in~\cref{alg:safe_opt}. It starts by computing the sets~$\mathcal{S}\ts{\ti}, \mathcal{G}\ts{\ti} \textnormal{ and } \mathcal{M}\ts{\ti}$ in~\cref{alg:safe_set,alg:expander_set,alg:maximizer_set}. Afterwards, a new evaluation point is chosen from the sets~$\mathcal{M}\ts{\ti}$ and~$\mathcal{G}\ts{\ti}$ in~\cref{alg:acquisition}, and the real system is evaluated in~\cref{alg:evaluate}. Finally, the GP is updated with the new, noisy measurement in~\cref{alg:update_gp}. This process is repeated until either the algorithm is aborted by the user or until a desired confidence, defined by~${\max_{\mb{a} \in \mathcal{G}\ts{\ti} \cup \mathcal{M}\ts{\ti}} w\ts{\ti}(\mb{a})}$, is reached~\cite{Sui2015Safe}.

Computing the complete set~$\mathcal{G}\ts{\ti}$ in~\cref{math:set_expanders} is computationally expensive, since we have to recompute the matrix inverse in~\cref{math:gp_prediction_mean,math:gp_prediction_variance} for every point in~$\mathcal{S}\ts{\ti}$.
However, since~\cref{alg:safe_opt} only selects the most uncertain parameter in~\cref{alg:acquisition}, it suffices to find the expander in~${\mathcal{S}\ts{\ti} \setminus \mathcal{M}\ts{\ti}}$ with the largest value~$w\ts{\ti}$ above the maximum variance in~$\mathcal{M}\ts{\ti}$, ${\mathrm{max}_{\mb{a} \in \mathcal{M}\ts{\ti}} w\ts{\ti}}$.
As a result, it suffices to iterate over the points in~${ \left\{ \mb{a} \in \mathcal{S}\ts{\ti} \setminus \mathcal{M}\ts{\ti} ~|~ w\ts{\ti}(\mb{a}) > \max_{\mb{a}' \in \mathcal{M}\ts{\ti}} w\ts{\ti}(\mb{a}') \right\} }$ in order of decreasing values~$w\ts{\ti}$ and stop the computation as soon as an expander is found.
This significantly reduces computation time, since typically only few or no parameters need to be checked as expanders using~\cref{math:expander_indicator}.

It is possible to extend this algorithm to additional constraints that do not depend on the performance, such as constraints on inputs or states. Please refer to~\cite{Berkenkamp2016Bayesian} for details.


\section{QUADROTOR EXPERIMENTS}
\label{sec:results}

In this section, we demonstrate the algorithm on a quadrotor vehicle, a Parrot AR.Drone 2.0.
A video of the experiments can be found at~\mbox{\url{http://tiny.cc/icra16_video}}.
The quadrotor learns optimal controller gains for the position controller in $x$-direction.
The other two directions and the heading angle are stabilized by separate controllers.
The system's dynamics can be described by four states: position,~$x$, velocity,~$\dot{x}$, pitch,~$\phi$, and angular velocity,~$\omega$.
Measurements of all states are available from an overhead motion capture camera system.
The control input,~$u$, is the desired pitch angle, which in turn is the input to an unknown, proprietary, on-board controller.
We define a linear control law, which computes the control input at time~$k$:
\begin{equation}
u\ts{k} = k_1 (x\ts{k} - r\ts{k}) + k_2 \dot{x}\ts{k}.
\label{math:linear_control_law}
\end{equation}
The control law depends on the reference position~$r\ts{k}$ and is parameterized by two parameters,~${ \mb{a} = (k_1, k_2) }$.

The goal is to find controller parameters that maximize the performance during a 1-meter reference position change. For an experiment with parameters~$\mb{a}_\ti$ at iteration~$\ti$,
\begin{align}
J(\mb{a}_\ti) &= C(\mb{a}_\ti) - 0.95 C(\mb{a}_0),
\label{math:performance_function} \\
C( \mb{a}_\ti ) &= -\sum_{k=0}^N \mathbf{x}_k^\mathrm{T} \mathbf{Q} \mathbf{x}_k + R u_k^2,
\label{math:cost_function}
\end{align}
where, to compute the cost~$C$, the states~${\mb{x} = (x - r, \dot{x}, \phi, \omega)}$ and the input~$u$ are weighted by positive semi-definite matrices~$\mathbf{Q}$ and~$R$.
The time horizon is~\unit[5]{s}~(${N=350}$).
Here we have defined performance as the cost improvement relative to $95\%$ of the initial controller cost. The safe threshold is set at ${ J_{\mathrm{min}}=0 }$.
In practice, ${J_{\mathrm{min}}}$ can be chosen freely; however, we cannot set the threshold equal to the performance of the initial controller, as this does not allow the algorithm to classify nearby states as safe and expand the safe set.

While the optimal controller gains could be easily computed given an accurate model of the system, we do not have a model of the dynamics of the proprietary, on-board controller and the time delays in the system.
Moreover, we want to optimize the performance for the real, nonlinear quadrotor system, which is difficult to model accurately.
An inaccurate model of the system could be used to improve the prior GP model of the performance function, with the goal of achieving faster convergence. In this case, the uncertainty in the GP model of the performance function would account for inaccuracies in the system model.

We discretize the controller parameter space uniformly into~$10,000$~combinations in~$[-0.6, 0.1]^2$, explicitly including positive controller parameters that certainly lead to crashes.
In practice, one would exclude parameters that are known to be unsafe~\textit{a priori}.
The initial controller gains are $(-0.4, -0.4)$, which result in a controller with poor performance.
Decreasing the controller gains further leads to unstable controllers.

To run the optimization algorithm we need to define a kernel for the GP. In this work, we choose the Mat{\`e}rn kernel with parameter~${\nu = 3/2}$~\cite{Rasmussen2006Gaussian},
\begin{align}
k(\mb{a}_i, \mb{a}_j) &= \sigma_\eta^2 \left( \hspace{-0.25em} 1 \hspace{-0.15em} + \hspace{-0.2em} \sqrt{3}\,r(\mb{a}_i, \mb{a}_j) \hspace{-0.25em} \right) \exp \left(\hspace{-0.25em}- \hspace{-0.05em} \sqrt{3} \,r(\mb{a}_i, \mb{a}_j) \hspace{-0.25em} \right) \hspace{-0.25em},
\label{math:matern_kernel} \\
r(\mb{a}_i, \mb{a}_j) &=  \sqrt{ (\mb{a}_i - \mb{a}_j)^\T \mb{M}^{-2} (\mb{a}_i - \mb{a}_j) },
\end{align}
which is parameterized by three hyperparameters: measurement noise,~$\sigma_\omega^2$ in~\cref{math:gp_prediction_mean,math:gp_prediction_variance}, prior variance,~$\sigma_\eta^2$, and postitive length-scales,~${\mb{l} \in \mathbb{R}_+^{|\mathcal{A}|}}$, which are the diagonal elements of the diagonal matrix~$\mb{M}$,
$\mb{M} = \mathrm{diag}(\mb{l})$, and correspond to the rate of change of the function~$J$ with respect to~$\mb{a}$.
This kernel function implies that the underlying function~$J$ is differentiable, takes values within the $2\sigma$ confidence interval~$[-2\sigma_\eta, 2\sigma_\eta]$ with high probability and, with high probability, has a Lipschitz constant that depends on~$\mb{l}$ and~$\sigma_\eta^2$.
These hyperparameters encode our prior assumptions about the unknown performance function.
While it may be difficult to find hyperparameters that describe the underlying function perfectly, it is usually possible to specify parameters that are conservative (i.e., large $~\sigma_\omega^2$ and~$\sigma_\eta^2$, and small length-scales,~$\mb{l}$).
In general, the more assumptions can be made (e.g., smoothness), the faster the algorithm will converge. This leads to a trade-off between ensuring that the function is well modeled (that is, the safe threshold is not violated), and the number of experiments we are willing to conduct.

\begin{figure}[t]
\vspace{1.5mm}
\centering
\includegraphics[scale=1]{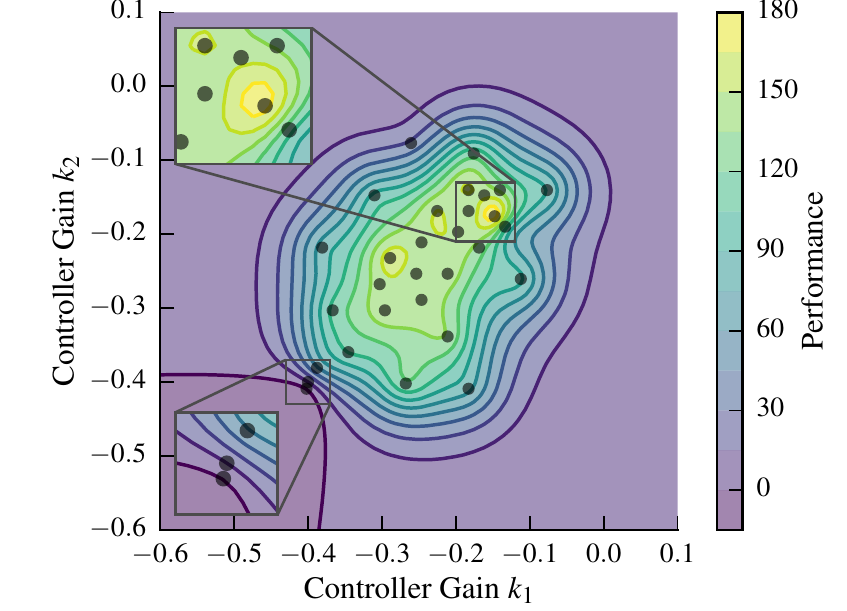}
\vspace{-0.6cm}
\caption{GP mean estimate of the performance function after~$30$ evaluations. The algorithm adaptively decides where to sample based on safety and informativeness. In the bottom-left corner, there is the magnified section of the first three samples, which are close together to determine the location of the initial, safe region. The maximum, magnified in the top-left corner, also has more samples to determine the precise location of the maximum. Other areas are more coarsely sampled to expand the safe region.}
\label{fig:final_performance_function}
\vspace{-0.1cm}
\end{figure}

The parameters for the experiments were set as follows: the length-scales were set to~$0.05$ for both parameters, which corresponds to the notion that a~0.05-0.1 change in the parameters leads to very different performance values. The prior standard deviation,~$\sigma_\eta$, and the noise standard deviation,~$\sigma_\omega$, are set to~$5\%$ and~$10\%$ of the performance of the inital controller,~$C(\mb{a}_0)$, respectively. The noise standard deviation,~$\sigma_\omega$, mostly models errors due to initial position offsets, since state measurements have low noise. The size of these errors depends on the choice of the matrices~$\mb{Q}$ and~$R$. By choosing~$\sigma_\omega$ dependent on the initial performance, we account for the~$\mb{Q}$ and~$R$ dependency. Similarly,~$\sigma_\eta$ specifies the expected size of the performance function values. Initially, the best we can do is to set this quantity dependent on the initial performance and leave additional room for future, larger performance values.

The resulting, estimated performance function after running~\cref{alg:safe_opt} for~30~experiments is shown in Fig.~\ref{fig:final_performance_function}.
The unknown function has been reliably identified. Samples are spread out over the entire safe set, with more samples close to the maximum of the function and close to the initial controller parameters. No unsafe parameters below the threshold~${ J_{\mathrm{min}} = 0 }$ were evaluated on the real system.

\begin{figure}[t]
\vspace{1.5mm}
\centering
\includegraphics[scale=1]{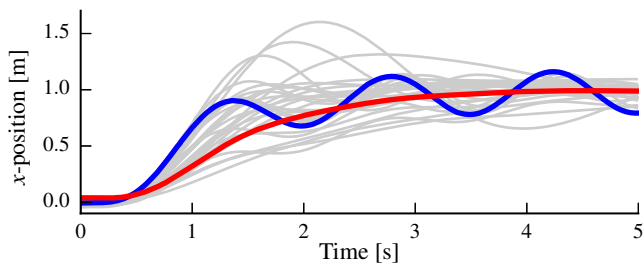}
\vspace{-0.3cm}
\caption{The quadrotor controller performance is evaluated during a~\unit[5]{s} evaluation interval, where a~\unit[1]{m} reference position change must be performed. The trajectories correspond to the optimization routine in~\cref{fig:final_performance_function}.
The initial controller (blue) performs poorly but is stable.
In contrast, the optimized controller (red) shows an optimized, smooth, and fast response.
The trajectories of other controller parameters that were evaluated are shown in gray.}
\label{fig:trajectories}
\vspace{-0.1cm}
\end{figure}

Typically, the optimization behavior of~\cref{alg:safe_opt} can be roughly separated into three stages.
Initially, the algorithm evaluates controller parameters close to the initial parameters in order for the GP to acquire information about the safe set (see lower-left, zoomed-in section in~\cref{fig:final_performance_function}).
Once a region of safe controller parameters is determined, the algorithm evaluates the performance function more coarsely in order to expand the safe set.
Eventually, the controller parameters are refined by evaluating high-performance parameters that are potential maximizers in a finer grid (see upper-left, zoomed-in section in~\cref{fig:final_performance_function}).
The trajectories of the initial, best and intermediate controllers can be seen in~\cref{fig:trajectories}.

A normal Bayesian optimization algorithm~\cite{Srinivas2010Gaussian} would start by evaluating ${ \mb{a} = (0.1,~0.1) }$, where the GP has the largest uncertainty about the performance.
These parameters lead to an unstable controller.
In contrast, our method safely explores the parameter space without evaluating unsafe parameters.

Ultimately, the algorithm identifies the controller gains that maximize the performance measure.
Because we omitted the on-board controller and its internal states and due to the nonlinearity of the quadrotor dynamics, the resulting performance function in Fig.~\ref{fig:final_performance_function} is similar to, but not the same as, the quadratic function that one would have expected from linear quadratic control theory.

\section{CONCLUSION}
\label{sec:conclusion}

We presented the first application of Safe Bayesian Optimization on a real robotic system. We modified the algorithm in~\cite{Sui2015Safe} to work directly with GP estimates and successfully applied it to the position control of a quadrotor vehicle.
It was shown that the algorithm enables efficient, automatic, and global optimization of the controller parameters without risking dangerous and expensive system failures.

\bibliographystyle{myIEEEtran}
\bibliography{root.bib}

\end{document}